\documentclass[11pt,a4paper]{article}
\usepackage[hyperref]{acl2019}
\usepackage{times}
\usepackage{latexsym}
\usepackage{ulem}
\usepackage{mathtools}
\usepackage{amsmath}
\usepackage{amssymb}
\usepackage{array}
\usepackage{tabu}
\usepackage{booktabs} 
\usepackage{multicol}
\usepackage{makecell}

\usepackage[utf8]{inputenc}

\usepackage{url}
\usepackage[colorinlistoftodos]{todonotes}

\PassOptionsToPackage{hyphens}{url}\usepackage{hyperref}

\aclfinalcopy 
\def\aclpaperid{57} 

\DeclareMathOperator*{\argmax}{argmax}

\title{Interrogating the Explanatory Power of Attention in Neural Machine Translation}

\author{Pooya Moradi, Nishant Kambhatla, and Anoop Sarkar \\
        Simon Fraser University \\
        8888 University Drive \\
        Burnaby, BC, Canada \\ 
        \texttt{\{pooya\_moradi, nkambhat, anoop\}@sfu.ca}}
\date{}

\begin{document}
\maketitle

\begin{abstract}
  Attention models have become a crucial component in neural machine translation (NMT). 
  They are often implicitly or explicitly used to justify the model's decision in generating a specific token but  
  it has not yet been rigorously established to what extent attention is a reliable source of information in NMT.
  To evaluate the explanatory power of attention for NMT, we examine the possibility of yielding the same prediction but with counterfactual attention models that modify crucial aspects of the trained attention model.
  Using these counterfactual attention mechanisms we assess the extent to which they still preserve the generation of function and content words in the translation process. 
  Compared to a state of the art attention model, our counterfactual attention models produce 68\% of function words and 21\% of content words in our German-English dataset.
  Our experiments demonstrate that attention models by themselves cannot reliably explain the decisions made by a NMT model.
  \footnote{The source code to reproduce the experiments is available at: \url{https://github.com/sfu-natlang/attention_explanation}}
\end{abstract}

\section{Introduction}

\par One shortcoming of neural machine translation (NMT), and neural models in general, is that it is often difficult for humans to comprehend the reasons why the model is making predictions \cite{feng2018,ghorbani2019}. The main cause of such a difficulty is that in neural models, information is implicitly represented by real-valued vectors, and conceptual interpretation of these vectors remains a challenge. 
Why do we want neural models to be interpretable? 
In order to debug a neural model during the error analysis process in research experiments, it is necessary to know how much each part of the model is contributing to the error in the prediction. Being able to interpret the deficiencies of a model is also crucial to further improve upon it. This requires an explainable understanding of the internals of the model, including how certain concepts are being modeled or represented. Therefore developing methods to interpret and understand neural models is an important research goal.
 
\begin{figure*}[t]
    \centering
    \includegraphics[width=1\textwidth]{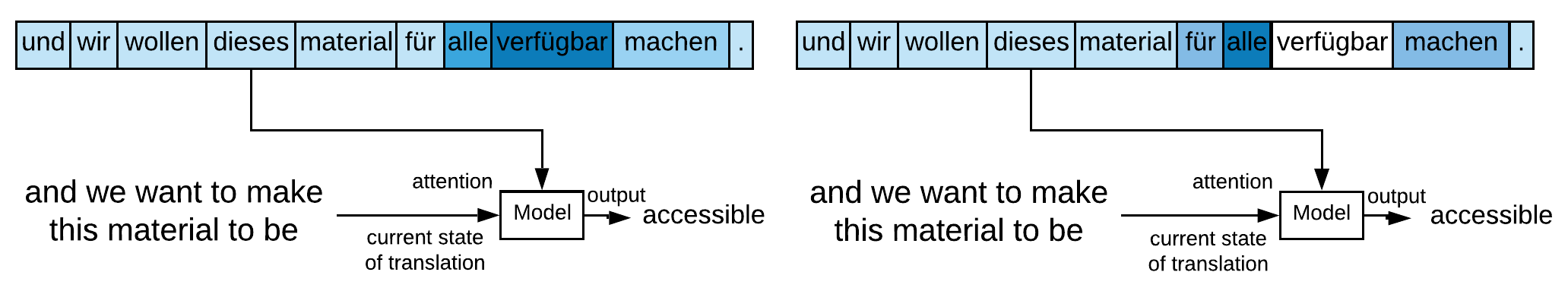}
    \caption{Two distinct attention weights yielding the same prediction. The model is translating the source word ``verfügbar'' to ``accessible''. In the left attention heatmap, the focus of the attention is on the word ``verfügbar''. However, in the right heatmap, ``verfügbar'' is not attended to at all and ``alle'' has received the most attention.
    }
    \label{fig:example}
\end{figure*}

\par
Visualizing and interpreting neural models has been extensively studied in computer vision \cite{simonyan2014,bach2015,zeiler2014,montavon2017}, and more recently in natural language processing (NLP) \cite{karpathy15,li2016,strobelt2017,strobelt2018}. Recently, the integration of attention mechanism \cite{bahdanau14} with an NMT sequence to sequence model~\cite{sutskever2014} has led to significant improvements in translation quality especially for longer sentence lengths. 
The attention mechanism provides a weighted average over the information from the source encodings to be used at each translation step. These weights are often regarded as a measure of importance, and are implicitly or explicitly used as an explanation for the model's decision. However it has not yet been established to what extent such explanations are reliable. 

The example in Figure \ref{fig:example} shows a model translating the German sentence ``und wir wollen dieses material für alle verfügbar machen.'' (\textit{and we want to make this material accessible to everyone.}). At the time that the model is translating ``verfügbar'' to ``accessible'', it is mostly attending to ``verfügbar'' (left heatmap). It is tempting to conclude that ``verfügbar'' having the most attention is why the model is generating the token ``accessible''. However, we manipulate the attention weights such that the ``verfügbar'' receives no attention and ``alle'' is given the most attention (right heatmap). We observe that in the second case, the model makes the same decision again and outputs ``accessible''. This example shows that using attention weights to reason about the model's predictions can be misleading, as these two heatmaps convey different explanations.

There are relatively few previous studies on investigating the power of attention mechanism in rationalizing a model's predictions in NLP \cite{jain2019-4,serrano2019} and they all target text classification tasks where attention is over the input document. Their findings do not generalize easily to NMT due to the difference in how the decoder works in the translation task which produces a sequence rather than a class label. 
NMT is a sequence-to-sequence (Seq2Seq) task, which is different from text classification. Also, the size of the output space is quite limited in text classification, whereas in NMT, it is equal to the vocabulary size of the target language that can be very large. Furthermore, different neural architectures (e.g., presence of an encoder and decoder in NMT), require different analysis and interpretations. To the best of our knowledge there is no existing work on addressing the question of interpretability of attention models for the machine translation task.


\par To investigate if the explanation implied by attention weights faithfully justifies why a model produced its output, we study to what extent it is possible to yield the same prediction by using counterfactual attention weights that suggest a contradictory explanation. The intuition behind this approach is that multiple contradictory justifications should not exist for a decision. To be specific, in the setting of an encoder-decoder model (Sec \ref{sec:encdec}), we propose several counterfactual attention weighting methods that suggest different explanations compared to the original attention weights (Sec \ref{sec:approach}) and analyze their performance in preserving the generation of function\footnote{The reference for function words (we added new function words including the EOS token to this) can be found at: \url{semanticsimilarity.files.wordpress.com/2013/08/jim-oshea-fwlist-277.pdf} } and content words (Sec \ref{sec:experiment}). Function words (e.g., \textit{a}, \textit{the}, \textit{is}) have little lexical meaning in contrast to content words and thus we are curious whether explanatory power of the attention mechanism differs for generation of these two groups of words.


\section{Encoder-Decoder Model with Attention Mechanism}
\label{sec:encdec}
Given a training sentence pair ($\mathbf{x}$, $\mathbf{y}$) where $\mathbf{x} = [x_1, x_2, ..., x_m]$ is a sentence in the source language and $\mathbf{y} = [y_1, y_2, ..., y_n]$ is its corresponding translation in the target language, the encoder, which is a recurrent neural network (RNN), runs over the source sentence to calculate the contextualized representation of the source words. Here, we use a bidirectional encoder, and concatenate the forward and backward hidden states to build the final representation.

\begin{equation}
\begin{aligned}
& \overrightarrow{h_t} = \overrightarrow{f_{enc}}(x_t, \overrightarrow{h_{t-1}}) \\
& \overleftarrow{h_t} = \overleftarrow{f_{enc}}(x_t, \overleftarrow{h_{t+1}}) \\
& h_t = [\overrightarrow{h_t},\overleftarrow{h_t}]
\end{aligned}
\end{equation}

Then the decoder starts to generate output tokens using the following probability distribution:
\begin{flalign*}
p(y_t|y_{<t}, x) = \textit{softmax}(g_{dec}(s_t, c_t))
\end{flalign*}

with $g_{dec}$ being a transformation function that produces a vocabulary-sized vector, and $s_t$ is the hidden unit of the decoder's RNN updated as:
\begin{flalign*}
s_t = f_{dec}(y_{t-1}, s_{t-1}, c_{t-1})
\end{flalign*}

where $f_{dec}$ is a RNN. Here $c_t$ is the context vector calculated by attention mechanisms:
\begin{flalign*}
c_t = \sum_{i=1}^{m}\alpha_{ti}h_i
\end{flalign*}

where $\alpha_{t}$ is the normalized attention weights over the source context:
\begin{flalign*}
\alpha_{ti}=\frac{e^{a(s_{t},h_i)}}
{\sum_{j}{e^{a(s_{t},h_j)}}}
\end{flalign*}

Here, $a$ is a scoring function that determines the contribution of each source context vector to the final context vector. Implementation of $a$ depends on the choice of the attention method. In this work, we use general attention \cite{luong2015} as the scoring function:
 \begin{flalign*}
 a(s_{t}, h_i) = s_{t}^\top W_ah_i
 \end{flalign*}

\section{Approach}
\label{sec:approach}
Given a trained NMT model $M$, a test sentence $\mathbf{x}$ with $\mathbf{y}$ being its translation generated by $M$ at the decoding step $t$ in which $\alpha_{t}$ is the attention vector, attending to the source word at position $m_t = \argmax_i \alpha_{t}[i]$ or $k$-best attended-to words in the source are often implicitly or explicitly regarded as a justification for the model's prediction at the time step $t$. 

A vital criteria for such a justification is that one should not be able to find a contradictory explanation for the model's decision. More precisely, if at the time of inference, it is possible to manipulate the original attention weights to constitute an alternative attention vector $\alpha^{\prime}_{t}$, such that $\argmax_i  \alpha^{\prime}_{t}[i]$ $\neq$ $m_t$  and the decision of the model is preserved, then these weights cannot be used for justification as they are contradictory. Thus, we are interested in assessing for what percentage of the words in the translation, counterfactual attention weights exist. These percentages can shed light on the reliability of the attention weights as a potential explanation. Note that at the inference time we manipulate attention vector for each output token separately and in isolation to make sure the output tokens at time steps $t+1$ and after will not be affected by the change at time step \textit{t}. This means that our output translations are unaffected by our counterfactual attention models which are purely for the examination of how attention might explain the model's decision.

The main task here is to find a counterfactual attention vector if it exists at all. An exhaustive search approach in which every possible attention vector is examined is computationally intractable and unlikely to provide much insight. Instead we experiment with a few specific counterfactual attention weighting methods that we think provide the most insight. It is important to note that in this case, the calculated percentage of the preserved words will be a lower-bound estimation for the true percentage. As a result, the explanation offered by attention weights are more unreliable than what we'll find. We experiment with the following attention methods to create counterfactual attention weights:

\begin{itemize}
    \item \texttt{RandomPermute} \cite{jain2019-4}: We set $\alpha^{\prime}_t = random\_permute(\alpha_t)$ such that $\argmax_i \alpha^{\prime}_{t}[i] \neq m_t$.
    
    \item \texttt{Uniform}: In this method, $\alpha^{\prime}_t = \frac{1}{m}\vec{1}$. Here, $m$ is the length of the source sentence.
    
    \item \texttt{ZeroOutMax}: A simple approach to create a counterfactual attention vector is to remove the maximum attention weight. So we set $a(s_{t}, h_{m_t})$ = $-\infty$.
\end{itemize}

We also experiment with four additional attention methods where our motivation is not finding counterfactual attention weights, but to improve our understanding of how attention weights influence the model's prediction:

\begin{itemize}
    \item \texttt{ZeroOut}: 
    In what conditions does the decoder overlook the information provided by attention mechanism? To answer this question, we set all attention weights to zero at inference time (attention is still used while training the model).
    
    \item \texttt{LastEncoderState}: Here, we only use the final hidden state of the encoder as the context vector to be used in the decoder. Note that this is different from seq2seq without attention in which final hidden state of the encoder is used to initialize the decoder.
    When the focus is on the final hidden state of the encoder in the original attention weights, this method does not produce a counterfactual attention vector, which is why we don't intend to use this method to create a contradictory explanation, but rather to gain more insight into the sensitivity of the model to attention weights.
    
    \item \texttt{OnlyMax}: In this and the following method, the source hidden state with the maximum attention still receives the highest attention, and so these two methods do not output counterfactual attention vectors. However we are curious to what extent other attention weights are required to preserve the model's output. Note that the weights produced by these methods can be counted as contradictory when multiple attention weights are used for justifying predictions. Because although the most attended source context is not changed, the relative ranking of the rest of the source context in terms of attention weights is changed. However, these kinds of justifications are mostly discussed in text classification. In this specific method we only keep the most attended source hidden state: $\alpha^{\prime}_t[m_t]$=1. 
    
    \item \texttt{KeepMaxUniformOthers}: Here, we set $\alpha^{\prime}_t[m_t]$ = $\alpha_{t}[m_t]$, but for all other positions $\alpha^{\prime}_t[i]$ = $(1 - \alpha^{\prime}_t[m_t]) / m$. This is to investigate if using other source hidden states uniformly has any added benefit.
    
\end{itemize}

\section{Experiments}
\label{sec:experiment}

\subsection{Data}
In this work we use the German-English dataset from IWSLT2014\footnote{
\url{https://sites.google.com/site/iwsltevaluation2014/}}. We concatenate dev2010, dev2012, tst2010, tst2011 and tst2012 to be used as the test data. Data is tokenized using Moses \cite{koehn2007}.

\subsection{Model Details}
OpenNMT \cite{klein2017} is used for our Seq2Seq implementation. We use Long Short-Term Memory (LSTM) as RNN units. Each LSTM unit has 2 layers, and the dimension size for LSTM units and word embeddings is set to 500. The model is trained using Adam trainer with learning rate 0.001 for 50000 steps using early stopping. Vocabulary size for both the source and target language is set to 50000. Sentences longer that 50 tokens are pruned.

\section{Results}
Table \ref{translation_statistics} shows the percentage of function and content words generated by the trained model. As expected, the majority of the generated tokens are function words. We discuss our findings in more detail in the subsequent subsections.

\begin{table}[!htbp]
\centering
\small{
\begin{tabular}{|l|c|}
\hline
Number of tokens (+EOS) & 139465 \\
 \hline
 Percentage of function words  & 68\%  \\
 \hline
 Percentage of content words  & 32\%  \\
\hline
\end{tabular} 
}
\caption{Percentage of function and content words in the generated translation.}
\label{translation_statistics}
\end{table}

\begin{table}[!htbp]
\centering
\small{
\begin{tabular}{|c|c|c|c|}
\hline
& \textit{Method} & \textit{\% for FWs} & \textit{\% for CWs} \\
 \hline
 1 & RandomPermute  & 33\% & 6\%  \\
 
 2 & Uniform & 53\% & 11\% \\
 
 3 & ZeroOutMax & 52\% & 15\% \\
 
 \hline
 4 & Aggregate(1+2+3) & \textbf{68\%} & \textbf{21\%} \\
 \hline\hline
 5 & ZeroOut & 9\% & 0\% \\
 6 & LastEncoderState & 20\% & 2\% \\
 7 & OnlyMax & 71\% & 83\% \\
 8 & KeepMaxUniformOthers & 86\% & 86\% \\
 
\hline
\end{tabular} 
}
\caption{\small{Percentage of the preserved function and content words in the proposed attention methods: Trying out all the methods to find a counterfactual attention vector maximizes the chance of success. We use methods in row 5-8 only to shed light on the sensitivity of the model's output to perturbation in attention weight. They are not necessarily counted as counterfactual attention methods. Higher preservation rate stands for better performance}.
}
\label{table:contradictory_attention_methods}
\end{table}

\subsection{Effectiveness of the proposed counterfactual attention methods}
Table \ref{table:contradictory_attention_methods} shows the percentage of function and content words for which counterfactual attention weights were found using the proposed attention methods. The \texttt{Uniform} method (row 2) is the most effective method to create counterfactual attention weights for function words. However, for content words, the \texttt{ZeroOutMax} method (row 3) is the most successful method. 

From Table \ref{table:contradictory_attention_methods}, we also derive that \texttt{RandomPermute} is not as effective as the \texttt{Uniform} and \texttt{ZeroOutMax} methods. Our justification is that in the \texttt{RandomPermute} method, it is highly probable that the context vector is biased toward a random source hidden state. Such bias can lead to misleading noise in the context vector. However, there isn't such a bias in the \texttt{Uniform} or \texttt{ZeroOutMax} methods. 

To maximize the chance of finding a counterfactual attention, for each output token, we try out all the proposed methods to check if we can find a counterfactual attention (row 4). As evident from Table \ref{table:contradictory_attention_methods}, this approach greatly increases the chance of finding a counterfactual attention. Note that as previously stated, these percentages are a lower-bound for the true percentage.

\subsection{Function words are more easily generated compared to content words}
An important observation in Table \ref{table:contradictory_attention_methods} is that the proposed methods are considerably more effective in preserving function words compared to content words. The production of function words rely more on the target context, in contrast to content words which rely more on the source context. Accordingly, perturbation in the original attention weights likely has significantly more impact on diminishing content words compared to function words. 

This ties well with the main idea behind context gates in which the influence of source context and target context is controlled dynamically \cite{tu2017}. Since the generation of function words relies more on the target context, one may wonder to what extent attention is needed for preserving function words? To answer this question, we completely zero out the attention using \texttt{ZeroOut}. Row 5 shows that only 9\% of function words were preserved in this method. 

Moreover it can be seen that the model could not preserve any content word when this method is employed. Interestingly, we found that the preserved function words in this method were all occurrences of ``,''. Apparently the decoder's language model is so strong in predicting ``,'' without attention. This finding suggests that a basic representation of the source context is still necessary to preserve function words.

\begin{table}[!tbp]
\centering

\begin{tabular}{|c|c|c|c|c|c|c|c|}
\hline
\textbf{Token} & \textbf{\# preserved} & \textbf{Coverage} \\
\hline
going & 310 & 70\% \\
people & 237 & 46\% \\
know & 219 & 62\% \\
world & 215 & 67\% \\
like & 189 & 47\% \\
think & 176 & 50\% \\
way & 162 & 68\% \\
get & 160 & 53\% \\
thing & 147 & 79\% \\
things & 142 & 56\% \\
time & 139 & 54\% \\
see & 137 & 51\% \\
years & 136 & 64\% \\
make & 126 & 49\% \\
little & 113 & 55\% \\
just & 109 & 29\% \\
really & 93 & 37\% \\
bit & 92 & 88\% \\
said & 89 & 59\% \\
got & 86 & 59\% \\
\hline

\end{tabular} 
\caption{Top 20 content words preserved by the aggregate method sorted by the number of times they were preserved.}
\label{table:aggregate_content_1}
\end{table}

\begin{table}[!htbp]
\centering

\begin{tabular}{|c|c|c|c|c|c|c|c|}
\hline
\textbf{Token} & \textbf{Coverage} & \textbf{Total} \\
\hline
bit & 88\% & 105 \\
course & 87\% & 91 \\
thank & 83\% & 89 \\
thing & 79\% & 186 \\
fact & 78\% & 74 \\
half & 78\% & 27 \\
own & 75\% & 75 \\
ones & 73\% & 30 \\
states & 73\% & 30 \\
difference & 71\% & 21 \\
going & 70\% & 444 \\
turns & 69\% & 26 \\
way & 68\% & 237 \\
able & 67\% & 85 \\
world & 67\% & 323 \\
doing & 66\% & 103 \\
planet & 65\% & 37 \\
years & 64\% & 212 \\
know & 62\% & 353 \\
united & 62\% & 21 \\
\hline

\end{tabular} 
\caption{Top 20 content words preserved by the aggregate method sorted by percentage of their total occurrences that are preserved (\textit{coverage}).}
\label{table:aggregate_content_2}
\end{table}

\subsection{Highlighting top preserved tokens}
An important question that may arise is whether each attention method tends to preserve a specific group of words. To address this question, we listed the top preserved function words and content words for all the proposed methods. We observed that they mostly preserve the same group of words but with different percentages. As a result, we only list the top preserved tokens for the aggregate method. 

Table \ref{table:aggregate_content_1} contains the top 20 content words sorted by the number of times they were preserved. It is interesting to note that for many of these frequent tokens, more than half of their total occurrences are preserved without focusing on their corresponding translation in the source sentence (e.g., ``going'', ``know'', ``thing'', etc). 

In Table \ref{table:aggregate_content_2}, we sort such tokens based on their coverage, which is the percentage of their total occurrences that are not affected when a counterfacual attention is applied\footnote{We consider only the tokens that have appeared more than 20 times. The reason is that there are many preserved words that have appeared only once (coverage=1) and it is not clear if the coverage remains the same when frequency increases.}. We repeat the same process for function words (Table \ref{table:aggregate_function_1} and Table \ref{table:aggregate_function_2}). 

As evident from Table 5, we have successfully yielded the same token in 94\% of the occurrences of the EOS token but with a contradictory explanation. This can be explained by the previous findings suggesting special hidden units keep track of translation length \cite{shi2016}. As a result, EOS token is generated upon receiving signal from these units rather than using attention. This indicates that attention weights are highly unreliable for explaining the generation of EOS tokens. This is worth noting because early generation of the EOS token is often a major reason of the under-translation problem in NMT \cite{kuang2018}. Thus, attention weights should not be used to debug early generation of EOS, and that some other underlying influence in the network \cite{ding2017} might be responsible for the model's decision in this case.

\begin{table}[!t]
\centering

\begin{tabular}{|c|c|c|c|c|c|c|c|}
\hline
\textbf{Token} & \textbf{\# preserved} & \textbf{Coverage} \\
\hline
, & 7329 & 85\% \\
EOS & 6364 & 94\% \\
the & 5210 & 82\% \\
. & 3947 & 60\% \\
of & 3003 & 87\% \\
to & 2923 & 86\% \\
and & 2639 & 67\% \\
a & 2187 & 65\% \\
that & 1936 & 69\% \\
i & 1737 & 76\% \\
\&apos;s & 1732 & 95\% \\
you & 1501 & 72\% \\
it & 1497 & 72\% \\
is & 1496 & 88\% \\
in & 1364 & 64\% \\
we & 1246 & 64\% \\
they & 624 & 69\% \\
\&quot; & 620 & 81\% \\
have & 613 & 70\% \\
be & 582 & 91\% \\
\&apos;t & 580 & 96\% \\
\&apos;re & 542 & 86\% \\
this & 541 & 42\% \\
so & 531 & 57\% \\
are & 526 & 77\% \\
was & 514 & 66\% \\
do & 433 & 77\% \\
about & 417 & 65\% \\
what & 415 & 61\% \\
can & 400 & 54\% \\
\hline

\end{tabular} 
\caption{Top 30 function words preserved by the aggregate method sorted by the number of times they were preserved.}
\label{table:aggregate_function_1}
\end{table}

\begin{table}[t]
\centering

\begin{tabular}{|c|c|c|c|c|c|c|c|}
\hline
\textbf{Token} & \textbf{Coverage} & \textbf{Total} \\
\hline
\&apos;t & 96\% & 602 \\
\&apos;s & 95\% & 1819 \\
EOS & 94\% & 6748 \\
be & 91\% & 641 \\
is & 88\% & 1707 \\
of & 87\% & 3450 \\
to & 86\% & 3383 \\
\&apos;re & 86\% & 631 \\
, & 85\% & 8582 \\
\&apos;m & 84\% & 311 \\
been & 82\% & 233 \\
lot & 82\% & 148 \\
the & 82\% & 6386 \\
\&quot; & 81\% & 770 \\
are & 77\% & 679 \\
do & 77\% & 565 \\
i & 76\% & 2290 \\
who & 73\% & 300 \\
it & 72\% & 2089 \\
you & 72\% & 2099 \\
have & 70\% & 876 \\
up & 70\% & 235 \\
they & 69\% & 904 \\
that & 69\% & 2812 \\
well & 67\% & 153 \\
and & 67\% & 3922 \\
was & 66\% & 774 \\
were & 65\% & 240 \\
same & 65\% & 154 \\
a & 65\% & 3369 \\

\hline

\end{tabular} 
\caption{Top 30 function words preserved by the aggregate method sorted by coverage.}
\label{table:aggregate_function_2}
\end{table}

\subsection{Last encoder hidden state is a poor representation of the source context in the attentional model}
\label{sec:remove_last_encoder_state}
The last encoder state in a non-attentional model is passed to the decoder as the representation of the source context \cite{cho2014} although the accuracy of this representation has been shown to degrade as sentence length increases \cite{bahdanau14}.
We experiment with the \texttt{LastEncoderState} method to investigate how well the last encoder state can be representative of the source context in the attentional setting, and if exclusively attending to it can be used as a counterfactual attention. 


Row 6 in Table \ref{table:contradictory_attention_methods} shows that there is a significant gap between the \texttt{LastEncoderState} and the counterfactual methods proposed in Table \ref{table:contradictory_attention_methods}. A possible explanation for this result is that in the presence of attention mechanism, the model is trained to distribute the source-side information among the encoder states to be easier to select the relevant parts of the information with respect to the decoding state. Consequently, the last encoder state does no longer capture the whole information.

If we look at Table \ref{last_encoder_state_top_10}, we can see that
the most covered functions words, are the words that usually appear at the end of the sentence (e.g., ``EOS'', ``.'', ``?'', ``!''). This is because most of the context captured by the last encoder state is centered around the last part of the sentence in which these tokens appear.

\subsection{Attention of non-maximum source hidden states}
\label{sec:multi_weight}
Row 7 shows that a majority of output tokens can be preserved even when the model attends to a single source hidden state. Row 8 shows that when other source hidden states are uniformly combined, although the ratio of unaffected content words has increased by 3\%, ratio of unaffected function words has increased by 15\%. This again underlines the importance of the basic representation of the source context for generation of function words. 

\begin{table}[!htbp]
\centering

\begin{tabular}{|c|c|c|c|}
\hline
\textbf{Token} & \textbf{Coverage} & \textbf{Total}  \\
 \hline

EOS & 99\% & 6748\\
. & 98\%  & 6526\\
? & 88\%  & 589 \\
\&apos;t & 86\%  & 602 \\
! & 68\%  & 22 \\
\&quot; & 60\%  & 770 \\
\&apos;s & 24\%  & 1819 \\
; & 24\%  & 63 \\
are & 18\%  & 679 \\
is & 18\% & 1707 \\

\hline
\end{tabular} 
\caption{Top 10 unaffected function words in the \texttt{LastEncoderState} method. }
\label{last_encoder_state_top_10}
\end{table}

\subsection{Summary}
Our findings can be summarized as follows:
\begin{itemize}
    \item It is possible to generate 68\% of function words and 21\% of content words with a counterfactual attention indicating unreliability of using attention weights as explanation.
    
    \item The generation of function words relies more on the target context, whereas the generation of content words relies more on the source context. This results in a higher likelihood of generation of preserved function words compared to that of preserved content words.
    
    
    \item Generation of EOS tokens cannot be reliably explained by using attention weights. Instead, this depends on the length of the target translation which is implicitly pursued by special hidden units. As a result, EOS token is emitted upon receiving a signal from these units rather than information from attention.

    \item The last encoder state is a poor representation of the source sentence and cannot be effectively used as the source context vector.
    
    \item It is possible to generate 86\% of tokens by only using the source hidden state with the maximum attention and using other source hidden states uniformly suggesting that it may not be necessary to assign highly tuned weights to each source hidden state.
    
\end{itemize}

\section{Related Work}
\label{sec:related_work}
Relevance-based interpretation is a common technique in analyzing predictions in neural models. In this method, inputs of a predictor are assigned a scalar value quantifying the importance of that particular input on the final decision. 
Saliency methods use the gradient of the inputs to define importance \cite{li2016,ghaeini2018,ding2019}. Layer-wise relevance propagation that assigns relevance to neurons based on their contribution to activation of higher-layer neurons is also investigated in NLP \cite{arras2016,ding2017,arras2017}. Another method to measure relevance is by removing the input, and tracking the difference in in network's output \cite{li2016b}. While these methods focus on explaining a model's decision,  \citet{shi2016,kadar2017,calvillo2018} investigate how a particular concept is represented in the network.

Analyzing and interpreting the attention mechanism in NLP  \cite{koehn2017,ghader2017,tang2018,clark2019,vig2019} is another direction that has drawn major interest. Although attention weights have been implicitly or explicitly used to explain a model's decisions, the reliability of this approach is not proven. Several attempts have been made to investigate the reliability of this approach for explaining a models' decision in NLP \cite{serrano2019, baan2019, jain2019-3, jain2019-4}, and also in information retrieval \cite{jain2019-2}. 

Our work was inspired by \citet{jain2019-4}. 
However, in this work we have focused on similar issues in neural machine translation which is has different challenges compared to text classification in terms of objective and architecture. Moreover, our paper studies the effect of different counterfactual attention methods.

\section{Conclusion}
\label{sec:conclusion}
Using attention weights to justify a model's prediction is tempting and seems intuitive at the first glance. It is, however, not clear whether attention can be employed for such purposes. There might exist alternative attention weights resulting in the same decision by the model but promoting different contradictory explanation. 

We propose several attention methods to create counterfactual attention weights from the original weights, and we measure to what extent these new contradictory weights can yield the same output as the original one. We find that in many cases, an output token can be generated even though a counterfactual attention is fed to the decoder. This implies that using attention weights to rationalize a model's decision is not a reliable approach. 

\section{Future Work}
In the future, we intend to study the extent to which attention weights correlate with importance measured by gradient-based methods. While we have separated function and content words in this work, we would like to extend our findings to other categories such as parts of speech (POS) or out-of-vocabulary (OOV) words. Another logical investigation for future would be to address interpretability of copy mechanism in NMT \cite{gu2016}.
Proving the correlation between attention and the model predictions in more sophisticated attention models such as Transformer \cite{vaswani2017} is also worth exploring.

\section*{Acknowledgments}

We would like to thank the anonymous reviewers for their helpful comments. The research was also partially supported by the Natural Sciences and Engineering Research Council of Canada grants NSERC RGPIN-2018-06437 and RGPAS-2018-522574 and a Department of National Defence (DND) and NSERC grant DGDND-2018-00025 to the second author.

\bibliography{acl2019}
\bibliographystyle{acl_natbib}

\end{document}